\documentclass[10pt,twocolumn,letterpaper]{article}

\usepackage[final]{cvpr}      % To produce the REVIEW version
\usepackage{graphicx}
\usepackage{amsmath}
\usepackage{amssymb}
\usepackage{booktabs}
\usepackage{amsfonts}
\usepackage{array}

\usepackage{algorithm}
\usepackage{algorithmic}

\usepackage{booktabs}
\usepackage{multirow}
\usepackage{color}
\usepackage{makecell}
\usepackage[table]{xcolor}         % colors

\usepackage{newfloat}
\usepackage{listings}
\definecolor{header}{RGB}{220,220,220}
\definecolor{subheader}{rgb}{0.63, 0.79, 0.95}
\newcommand{\Tstrut}{\rule{0pt}{2.6ex}}
\newcommand{\Bstrut}{\rule[-0.9ex]{0pt}{0pt}}
\newcommand{\TBstrut}{\Tstrut\Bstrut}

\usepackage[pagebackref,breaklinks,colorlinks]{hyperref}
\usepackage{amsthm} 
\usepackage{thmtools}

\usepackage[capitalize]{cleveref}
\usepackage{amsmath,amssymb,xpatch}

\crefname{section}{Sec.}{Secs.}
\Crefname{section}{Section}{Sections}
\Crefname{table}{Table}{Tables}
\crefname{table}{Tab.}{Tabs.}
  
\newtheoremstyle{exampstyle}
	{} % Space above
	{-2pt} % Space below
	{\itshape} % Body font
	{} % Indent amount
	{\bfseries} % Theorem head font
	{.} % Punctuation after theorem head
	{.5em} % Space after theorem head
	{} % Theorem head spec (can be left empty, meaning `normal')

\theoremstyle{exampstyle}

\newtheorem{theorem}{\textit{Theorem.}}
\newtheorem{lemma}{\textit{Lemma}}
\newtheorem{remark}{\textit{Remark}} 
\newtheorem{property}{\textit{Property}}

\def\cvprPaperID{*****} % *** Enter the CVPR Paper ID here
\def\confName{CVPR}
\def\confYear{2023}

\begin{document}

%%%%%%%%% TITLE - PLEASE UPDATE
\title{Stable and Efficient Adversarial Training through Local Linearization}

\author{Zhuorong Li, Dawei Yu\\
Zhejiang University City College\\
Hangzhou, China\\
{\tt\small lizr@zucc.edu.cn, ydw.ccm@gmail.com}}
% For a paper whose authors are all at the same institution,
% omit the following lines up until the closing ``}''.
% Additional authors and addresses can be added with ``\and'',
% just like the second author.
% To save space, use either the email address or home page, not both
%\and
%Daiwei Yu\\
%Zhejiang University City College\\
%First line of institution2 address\\
%{\tt\small ydw.ccm@gmail.com}
%}
\maketitle

%%%%%%%%% ABSTRACT
\begin{abstract}
		There has been a recent surge in single-step adversarial training as it shows robustness and efficiency. However, a phenomenon referred to as ``catastrophic overﬁtting" has been observed, which is prevalent in single-step defenses and may frustrate attempts to use FGSM adversarial training. To address this issue, we propose a novel method, \textit{\textbf{S}table and \textbf{E}fficient \textbf{A}dversarial \textbf{T}raining} (\textbf{\textit{SEAT}}), which mitigates catastrophic overfitting by harnessing on local properties that distinguish a robust model from that of a catastrophic overfitted model. The proposed SEAT has strong theoretical justifications, in that minimizing the SEAT loss can be shown to favour smooth empirical risk, thereby leading to robustness. Experimental results demonstrate that the proposed method successfully mitigates catastrophic overfitting, yielding superior performance amongst efﬁcient defenses. Our single-step method can reach 51\% robust accuracy for CIFAR-10 with $l_\infty$ perturbations of radius $8/255$ under a strong PGD-50 attack, matching the performance of a 10-step iterative adversarial training at merely 3\% computational cost.
\end{abstract}

%%%%%%%%% BODY TEXT
\section{Introduction}\label{sec:1}
Whereas Deep Neural Networks(DNNs) based systems penetrate almost every corner in our daily life, they are not intrinsically robust. In particular, by imposing high fidelity and imperceptible distortion on the original inputs, also known as adversarial attacks, decisions of DNNs can be completely altered \cite{ml-3,ml-2}. It is thus imperative to develop mitigation strategies, especially in high-stakes applications where DNNs are applied, \eg, autonomous driving and surveillance system \cite{ml-7,ml-8}.

There has been a great deal of work on devising sophisticate adversarial attacks \cite{sat-30,ml-9}, which has thus spurred immense interest towards building defenses against such attacks \cite{ml-13,ml-14,ml-15}. Among them, \textit{adversarial training} (AT) is one of the most promising methods to achieve empirical robustness against adversarial attacks. Its training regime attempts to directly augment the training set with adversarial samples that are generated on-the-fly \cite{sat-7,madry}. Specifically, when the adversarial samples above are produced by multiple gradient propagations, the corresponding adversarial training is named multi-step AT or otherwise single-step AT. Unfortunately, the cost of AT becomes prohibitively high with growing model capacity and dataset scale. This is primarily due to the intensive computation of adversarial perturbations, as each step of adversarial training requires multiple forward propagations to find the perturbations. 
\begin{figure}[t]
	\centering
	\includegraphics[width = \linewidth]{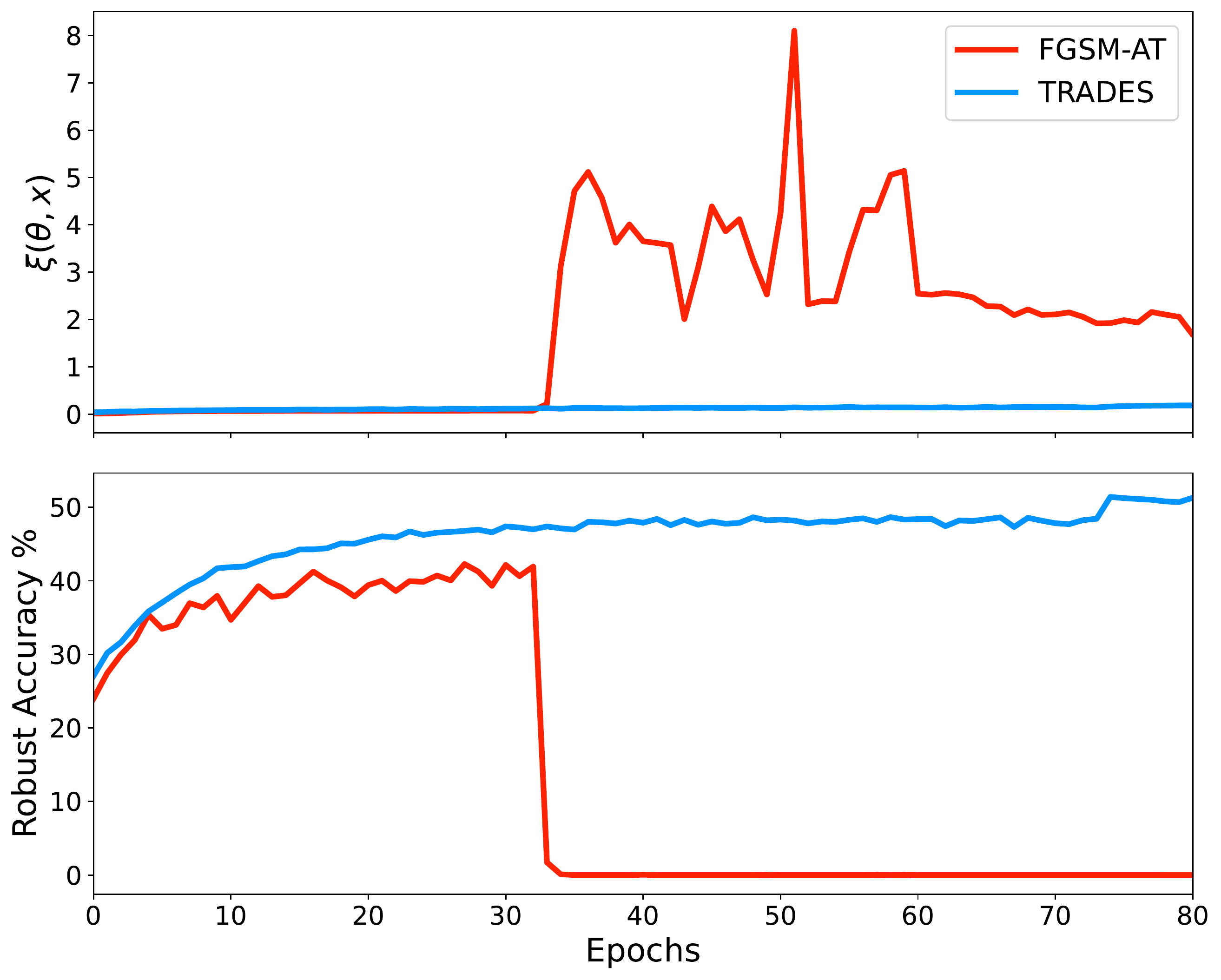}
	\caption{Analysis of catastrophic overfitting. Solid lines indicate the robust accuracy against PGD-20 on validation set during the training process, and dashed lines denote the mean linearity approximation error over entire training set.}
	\label{Fig:1}
\end{figure}

One approach to alleviate such computational cost is to train with single-step adversary, such as Fast Gradient Sign Method (FGSM) \cite{sat-7}, which takes only one gradient step to compute thus is much cheaper. However, this method relies on the local linear assumption of the loss surface, which is often compromised as the FGSM training progresses. As firstly discovered by \cite{fast}, single-step AT is prone to an intriguing phenomenon referred as \textit{catastrophic overfitting}, \textit{i.e.}, the validating robustness under multi-step attack, \textit{e.g.}, projected gradient descent (PGD) \cite{madry}, suddenly drops to zero whereas the training robustness against FGSM attack keeps increasing. Later, it was found that catastrophic overfitting it not limited to FGSM based AT but also occurs in diverse single-step AT methods \cite{gradalign}.  Few attempts \cite{sads,li,gradalign} have been made to identify the underlying reason for catastrophic overfitting and thus to develop strategies to prevent this failure. However, these work did not provide a fundamental reason for the problem and methods proposed are computationally inefficient \cite{sub}. 

In this work, we first unfold the connection behind catas-
trophic overfitting and local linearity. Recall that single-step adversary such as FGSM produces perturbations based on liner approximation of the loss function. However, the gradient masking causes the linear assumption to become unreliable, which results in the exclusion of strong attacks that maximize the loss during training. As catastrophic overfitting typically happens in non-iterative AT, we perform a comparison between single-step AT and multi-step AT for our empirical study. \cref{Fig:1} shows that catastrophic overfitting coincides with the drastic change of local linearity of the loss surface. To be more specific, the linearity approximation error of FGSM-AT abruptly increases in the moment when catastrophic overfitting occurs, with the robust accuracy suddenly deteriorated within an epoch. On the contrary, TRADES that generates adversarial examples with multiple iterations, is able to maintain negligible value of local linearity error during entire training. 

Upon this observation, we attempt to make a rigorous and comprehensive study on addressing catastrophic overfitting by retaining the local linearity of the loss function. The proposed \textbf{\textit{S}}\textit{table} and \textbf{\textit{E}}\textit{fficient} \textbf{\textit{A}}\textit{dversarial} \textbf{\textit{T}}\textit{raining} harnesses on the local linearity of models trained using multi-step methods, and incorporates such salient property into models trained using single-step AT. \cref{Fig:appendix} in Appendix shows that model resulting from the proposed SEAT behaves in a strikingly similar way to that of TRADES, indicating competitive robustness with multi-step AT.

Our main contributions are summarized as follows:
% \vspace{0.15cm}
% \begin{itemize}
% 	\item We first empirically identify a clear correlation between catastrophic overfitting and local linearity in DNNs, which sparks our theoretical analysis of the failure reason behind FGSM-AT, motivating us to overcome the shortcoming.
% 	\item We propose a novel regularization, \textbf{\textit{S}}\textit{table} and \textbf{\textit{E}}\textit{fficient} \textbf{\textit{A}}\textit{dversarial} \textbf{\textit{T}}\textit{raining} (\textit{\textbf{SEAT}}), which prevents catastrophic overfitting by explicitly punishing the violation of linearity assumption to assure the validity of FGSM solution.
% 	\item We conduct a throughout experimental study and show that the proposed SEAT consistently achieves superiority in stability and adversarial robustness amongst existing single-step AT, and is even comparable to most multi-step AT but at a much lower cost. We also justify the effectiveness of SEAT in different attack setups, and through loss surface smoothness, and decision boundary distortions.
% \end{itemize}

$\bullet$ We first empirically identify a clear correlation between catastrophic overfitting and local linearity in DNNs, which sparks our theoretical analysis of the failure reason behind FGSM-AT, motivating us to overcome the weakness.

$\bullet$ We propose a novel regularization, \textbf{\textit{S}}\textit{table} and \textbf{\textit{E}}\textit{fficient} \textbf{\textit{A}}\textit{dversarial} \textbf{\textit{T}}\textit{raining} (\textit{\textbf{SEAT}}), which prevents catastrophic overfitting by explicitly punishing the violation of linearity assumption to assure the validity of FGSM solution.

$\bullet$ We conduct a throughout experimental study and show that the proposed SEAT consistently achieves superiority in stability and adversarial robustness amongst existing single-step AT, and is even comparable to most multi-step AT but at a much lower cost. We also justify the effectiveness of SEAT in different attack setups, and through loss surface smoothness, and decision boundary distortions.

\section{Related Work}\label{sec:2}
\subsection{Adversarial Robustness and Attack Strength}\label{sec:2.1}
Adversarial training is widely regarded as the most eﬀective defenses. According to the count of gradient propagations involved in attack generation, methods can be mainly grouped into multi-step AT \cite{madry,awp-6} and single-step AT \cite{co,sat-7}. Multi-step AT, such as PGD-AT \cite{madry}, generally achieves robustness by training on strong perturbations generated by iterative optimization. In more recent work, TRADES \cite{trades} and AWP \cite{sub-40-awp} yield enhanced robustness with regard to a regularization, and \cite{pang} further improves by a judicious choice of hyperparameters. Albeit empirically the best performing method to train robust models, multi-step AT is time-consuming.

The high cost of multi-step AT has motivated an alternate, \textit{i.e.}, single-step AT, which proves to be efficient. It trains by training with shared gradient computations \cite{free}; or by using cheaper adversaries, such as FGSM \cite{sat-7,gat}; or by ﬁrst using FGSM and later switching to PGD \cite{convergence}. While these single-step AT shows promising direction, their robust performance is not on par with multi-step AT. Worse still, it is prone to a serious problem of \textit{catastrophic overfitting}, \textit{i.e.} after a few epochs of adversarial training, the robust accuracy of the model against PGD sharply decreases to 0\%. 
\subsection{Adversarial Generalization and Flat Minima}\label{sec:2.2}
There exists a large body of work investigating the correlation between the ﬂatness of local minima and the generalization performance of DNNs on natural samples \cite{amp-26,awp-22}. It has been empirically veriﬁed and commonly accepted that ﬂatter loss surface tends to yield better generalization, and this understanding is further utilized to design regularization(\textit{e.g.}, \cite{amp-10,amp-44,flood}).

Analogous connection has also be identified in adversarial training scenario, where the flatness of loss surface helps to improve robust generalization on adversarial samples \cite{awp-29}. Several well-recognized improvements of AT, \textit{i.e.}, TRADES \cite{trades}, MART \cite{mart}, and RST \cite{awp-6}, all implicitly flatten the loss surface to improve the robust generalization. Moreover, there are a line of works proposed explicit regularization to directly encourage the flatness of local minima \cite{llr, sub-40-awp}.

\section{Methodology}\label{sec:3}
Our aim is to develop a technique that resolves catastrophic overfitting so as to stabilize the single-step adversarial training. In this section, we ﬁrst theoretically analyze the pitfalls of present single-step AT methods. Then, we provide theoretical justiﬁcations for our regularization, proving our potentials to avoid catastrophic overﬁtting. Finally, we expound on our proposal, which is termed ``Stable and Efﬁcient Adversarial Training" or SEAT.  
\subsection{Revisiting Single-step Adversarial Training} \label{sec:3.1}
Recall the adversarial training, where adversarial perturbation $\delta$ can be generated by:
\begin{align}
	\delta \gets \text{Proj}\big(  \delta - \epsilon \cdot\nabla_\delta \ell(x+ \delta) \big)
	\label{eq:sec3.1}
\end{align}
where $\text{Proj}(x) = \arg \min _{\xi \in \mathcal{B}(r)} \Vert x - \xi\Vert_p$, $r$ is small value radius, $x \in \mathcal{D}$ is the sampled data that with ground truth label $y$, $\ell(\cdot)$ is the loss function, $\epsilon$ is an arbitrary perturbation with sufficiently small size, and $\nabla_\delta\ell(\cdot)$ denotes the gradient of the loss \textit{w.r.t.} perturbation $\delta$. While the loss function is highly non-linear, we can approximate it by making a reasonable assumption that the loss is once-differentiable. 

So for the adversarial loss $\ell(x+\delta)$, according to the first-order Taylor expansion, it leads to:
\begin{align}
	\ell({x+\delta}) = \ell(x) + \langle \delta, \nabla_x\ell(x) \rangle + \omega(\delta),
	\label{eq:sec3.2}
\end{align}
where $x+\delta \in \mathcal{B}_r(x)$ denotes a small neighborhood of the $x$. Suppose that given a metric space $M=(\mathcal{X},d)$,  there exists a radius  $r>0$ satisfying $\mathcal{B}_r(x)=\{ x+\delta\in \mathcal{X}: d(x,x+\delta)<r \}$. Besides, $\omega(\delta)$ denotes the higher order terms in $\delta$, which tends to be zero in the limit of a small perturbation. 

As we focus on single-step attacks in this work, we choose $\delta = \epsilon \cdot \text{sgn}(\nabla_x \ell(x))$ where $\epsilon$ is a small value and $\text{sgn}(\nabla_x\ell(x))$ is a vector. Now \cref{eq:sec3.2} would be, 
\begin{align*}
	\begin{split}
		\ell(x+\delta) &= \ell(x) + (\nabla_x \ell(x)^T) \epsilon \cdot \text{sgn}(\nabla_x\ell(x)) + \omega(\delta) \\
		&=\ell(x) + \epsilon\cdot \Vert \nabla_x\ell(x) \Vert_1 + \omega(\delta)
% 		\label{eq:sec3.3}
	\end{split}
\end{align*}
As the adversarial perturbation $\delta$ is carefully crafted to be imperceptible, it naturally satisfies that $\omega(\delta)$ is negligible. Thus, we have: 
\begin{align}
	\ell(x+\delta) - \ell(x) = \epsilon \cdot \Vert \nabla_x\ell(x) \Vert_1
	\label{eq:sec3.4}
\end{align}
As described in the previous section,  catastrophic overfitting has been observed during the training phase of models using single-step defenses. This indicates that the maximum magnitude $\epsilon$ is no longer the strongest step size in the direction of perturbation $\delta$. Subsequently, the maximization cannot be satisﬁed when catastrophic overfitting occurs as the loss surface is highly curved. \cref{eq:sec3.4} reveals that a fix step of perturbation, \textit{i.e.}, $\epsilon \cdot \Vert \nabla_x \ell(x) \Vert_1$ is probably the principle reason for catastrophic overfitting. 

\subsection{Justifications for Local Linearization}\label{sec:3.2}
To resolve the issue mentioned above, we suggest a regularizer to prevent catastrophic overﬁtting by encouraging local linearity, coupled with a scaled step size. 

As described before,  suppose that we are given a loss function $\ell(\cdot)$ that is once-differentiable, the loss at the point $x+\delta \in \mathcal{B}_r(x)$, \textit{i.e.}, the adversarial loss $\ell(x+\delta)$, can be well approximated by its first-order Taylor expansion at the point $x$ as $\ell(x) + \langle \delta, \nabla_x\ell(x) \rangle $. In other words, we can measure how linear the surface is within a neighborhood by computing the absolute difference between these two values as
\begin{align}
	\xi(\theta,x) = \vert \ell({x+\delta}) - \ell(x)-\langle\delta,\nabla_x\ell(x)\rangle \vert,
	\label{eq:sec3.5}
\end{align}
where the $\xi(\theta,x)$ defines \textit{linearity approximation error}, and the perturbation we focus on is generated by non-iterative adversary as  $\delta = \epsilon \cdot \text{sgn}(\nabla_x \ell(x))$.

Next, we provide theoretical analysis on how the linearity approximation error $\xi(\theta,x)$ correlates with the catastrophic overfitting, thereby deriving a regularizer for stabilizing the training phase.
\vspace{-0.1cm}
\begin{property}
	Consider a loss function $\ell(\cdot)$ that is once-differentiable, and a local neighbourhood $\mathcal{B}_\epsilon(x)$ that of radius $\epsilon$ and centered at $x$. For any $x+\delta \in \mathcal{B}_\epsilon(x)$, we have
	\vspace{-0.4cm}
	\begin{align}
		\begin{split}
			&\vert \ell({x+\delta}) - \ell(x)-\langle\delta,\nabla_x\ell(x)\rangle \vert \\ &\leq (\epsilon + \frac{1}{\sqrt{n}} \cdot\Vert \delta \Vert_2) \cdot \Vert \nabla_x\ell(x) \Vert_1.
			\label{eq:sec3.6}
		\end{split}
	\end{align}
	\label{property:1}
\end{property}
\vspace{-0.2cm}
\noindent \textbf{\textit{Proof.}} We start from the local Taylor expansion \cref{eq:sec3.2}. It is clear that when $x\to x+\delta$, the higher order term $\omega(\delta)$ becomes negligible. Mostly, we instead use its equivalent, \textit{i.e.}, $\vert \ell({x+\delta}) - \ell(x)-\langle\delta,\nabla_x\ell(x)\rangle \vert$ as a measure of how linear the surface is within a neighbourhood. Apparently, this is precisely the regularizer we defined in \cref{eq:sec3.5}.
\begin{align}
	\begin{split}
		&\vert \ell({x+\delta}) - \ell(x)-\langle\delta,\nabla_x\ell(x)\rangle \vert \\
		\leq &\vert  \ell({x+\delta}) - \ell(x) 
		\vert + \vert  \langle\delta,\nabla_x\ell(x)\rangle  \vert \\
		=&\epsilon\cdot \Vert \nabla\ell(x) \Vert_1  + \vert  \langle\delta,\nabla_x\ell(x)\rangle  \vert \\
		\leq &\epsilon\cdot \Vert \nabla\ell(x) \Vert_1  + \Vert \delta\Vert_2 \cdot \Vert \nabla_x\ell(x)\Vert_2 \\
		\leq &(\epsilon + \frac{1}{\sqrt{n}} \cdot \Vert \delta \Vert_2) \cdot \Vert \nabla\ell(x) \Vert_1
	\end{split}
	\label{eq:A3}
\end{align}

\begin{remark}
	Upon \cref{eq:sec3.6} it is clear that by bounding the linearity approximation error, one can implicitly introduces a scaling parameter $\Vert \delta \Vert_2 / \sqrt{n}$ for perturbation generation instead of a fixed magnitude $\epsilon \cdot \Vert \nabla_x \ell(x) \Vert_1$ of FGSM attack, which is the main cause of catastrophic overfitting.
\end{remark}
\vspace{-0.05cm}
\begin{property}
	Consider a loss function $\ell(\cdot)$ that is locally Lipschitz continuous. For any point in $\mathcal{B}_\epsilon(x)$, linearity approximation error $\xi(\theta,x)$ gives the following inequality:
	\vspace{-0.05cm}
	\begin{align*}
		\begin{split}
			K <\sup \limits_{x+\delta \in \mathcal{B}_\epsilon(x)} \dfrac{1}{\Vert \delta \Vert _2} \cdot &\Big( \vert \ell(x+\delta) - \ell(x)-\langle \delta,\nabla_x\ell(x)\rangle \vert  \\ &+ \vert \langle \delta,\nabla_x\ell(x) \rangle \vert \Big)		
		\end{split}
		% 		\label{eq:sec3.7}
	\end{align*}
	\label{property:2}
\end{property}
\noindent Details of the proof are provided in \cref{Appendix:B}.

\begin{remark}
	The \textit{linearity approximation error} $\xi(\theta, x)$ composes the expression on the RHS. Hence, minimizing $\xi(\theta, x)$ is expected to induce a smaller locally Lipschitz constant \textit{K}, thereby encouraging the optimization procedure to find a model that is locally Lipschitz continuous.
\end{remark}
Based on the above favorable bound~(as \textit{Property} \ref{property:1}) and local Lipschitz~(as \textit{Property} \ref{property:2}) continuity, we derive our proposed regularizer from the linearity approximation error
\begin{align}
	J(\theta) = \lambda \cdot \vert\max \limits_{\delta \in \mathcal{B}_\epsilon(x)}\ell(x+\delta)-\ell(x)-\delta^T\nabla_x \ell(x)\vert.
	\label{eq:sec3.8}
\end{align}
where $\lambda \in \mathbb{R}$ is a hyperparameter specifying the strength for imposing local linearization. We set $\lambda=0.5$ based on the experiment presented in \cref{tab:3}, and also for a well balance.

\subsection{Training Objective}\label{sec:3.3}
The proposed method incorporates the linearization regularizer to enhance optimization in both attack generation and defense training. In particular, we modify the training scheme of vanilla multi-step defense and propose our training objective of SEAT as follows:
\begin{align}
	&\min \limits_{\theta} \mathbb{E}_{(x,y)} \,\,  \widetilde{L}_{\text{adv}}\big(f_\theta(x+\delta^*),y\big) + J(\theta) \label{eq:9}
	\\
	&\delta^* = \arg \max \limits_{x+\delta \in \mathcal{B}_\epsilon(x)} L_{\text{adv}}\big(f_\theta(x+\delta),y\big) + J(\theta)
\end{align}
where $L_\text{adv}$ is the standard loss, \textit{e.g.}, the cross-entropy loss or maximum margin loss, $J(\theta)$ is the proposed  linearization regularizer and $f_\theta(\cdot)$ represents the neural network with parameters $\theta$. We will defer the deﬁnition of $ \widetilde{L}_\text{adv}$ later. Pseudo-code for our algorithm is given in \cref{al}.

\begin{algorithm}
	\caption{\textit{\textbf{S}table and \textbf{E}fficient \textbf{A}dversarial \textbf{T}raining}}
	\begin{algorithmic}
		\REQUIRE Total Epoch $N$, Neural Network $f_\theta$ with parameters $\theta$, Training Set $\mathcal{D} = \{ (x_j,y_j) \}$, Adversarial Perturbation Radius $\epsilon$, Step Size $\alpha$, Flooding level $b$.
		\FORALL{$\text{epoch}=1, \cdots, N$}
		\FORALL{$(x_j,y_j) \in \mathcal{D}$}
		\STATE {\color{blue}{* Inner Maximization to update $\delta$}}
		\STATE \quad $\delta\,=$\,\,Uniform\,$(-\epsilon, \epsilon)$
		\STATE \quad $L = L_{\text{adv}}(x_j + \delta,y_j) + J(\theta)$
		\STATE \quad \quad $=-f_\theta^{y_j}(
		x_j + \delta) + f_\theta^j(x_j+\delta) + J(\theta)$
		\STATE \quad $\delta^* \gets \delta + \alpha \cdot \text{sgn}(\nabla_\delta L)$
		\STATE {\color{blue}{* Outer Minimization to update $\theta$}}
		\STATE \quad $\widetilde{L} = \widetilde{L}_{adv} (x_j + \delta^*,y_j) + J(\theta)$
		\STATE \quad \quad $=\vert 
		L_{ce}(x_j + \delta^\ast, y_j) - b \vert + b + 
		J(\theta)$
		\STATE \quad $\theta \gets \theta - \eta \cdot (\nabla_\theta \widetilde{L})$
		\ENDFOR
		\ENDFOR
	\end{algorithmic}
	\label{al}
\end{algorithm}

\subsubsection{Outer Minimization}
Upon previous empirical observation and theoretical analysis, we assume that catastrophic overﬁtting is probably correlated with the deterioration of local linearity. To resolve this problem, we motivate SEAT through local linearization. The proposed training scheme caters to the dual objective of minimizing the classification loss on adversarial examples, while also explicitly minimizing the violation of linearity assumption to assure the validity of FGSM solution. Our preliminary is to use the cross-entropy loss and further introduce the proposed regularizer derived from the linearity approximation error.

As is observed that training models excessively towards adversarial robustness may hurt the generalization \cite{odds}, this work takes one more step towards mitigating overfitting. Based on the understand that flat minima tend to yield better generalization, we leverage the recently proposed regularization Flooding \cite{flood}, which forces the training loss to stay above a reasonably small values rather than approaching zero to avoid overfitting. To the best of our knowledge, we are the first ever to introduce the Flooding into single-step adversarial training, motivating to produce models that can be better generalized to multi-step optimized attacks,  thereby mitigating the catastrophic overfitting of the same. 

The implementation of Flooding is surprisingly simple as $\widetilde{R}(\theta) = \vert {R}(\theta) - b \vert + b$, where ${R}(\theta)$ and $\widetilde{R}(\theta)$ respectively denotes the original training objective and the flooded training objective, and $b > 0$ is the flood level \cite{flood}. Thus, we are inspired to use the flooded version instead of the basic adversarial training loss. 

Building on it, we replace the adversarial loss $L_{\text{adv}}$ with $\widetilde L_\text{adv}$. Now, we arrive at our implementation of the outer minimization for our single-step defense SAET, as \cref{eq:9}.

\subsubsection{Inner Maximization}
On the attackers’ side, they are trying to find the perturbation in which not only the loss on adversarial samples is maximized, but also the linearity assumption in the vicinity of each data point is maximally violated. Further, we use the maximum margin loss for the implementation of $L_{\text{adv}}$, as it is known to be beneficial especially for single-step AT, which heavily relies on the initial gradient direction \cite{marg,gat}. The maximum margin loss can be given by $-f_\theta^y(\widetilde{x})+\max f_\theta^j (\widetilde{x})$, where $\widetilde{x}$ denotes the adversarial image, $f_\theta^y(\widetilde{x})$ is the score on $\widetilde{x}$ with ground truth $y$ and $j\neq y$. As the ablation study in \cref{Appendix:F} shown, the maximum margin loss, coupled with our proposed linearization regularizer, improves the attack efficacy and thereby yielding models that are signiﬁcantly more robust.

\begin{table*}[h]
	\centering
	\caption{\textbf{White-box evaluation (CIFAR-10).} Performance (\%) of different defenses under various adversaries. Methods presented here for comparison are as follows: FGSM-AT \cite{sat-7}, RS-FGSM \cite{fast}, SubAT \cite{sub}, GradAlign \cite{gradalign}, SSAT \cite{co}, GAT\cite{gat}, PGD \cite{madry} and TRADES \cite{trades}.}
	\begin{tabular}{c c c c c c c c c c c}
		\toprule[1.5pt]
		\multirow{2}{*}{} & \multirow{2}{*}{Method}    & \multirow{2}{*}{Steps} & \multirow{2}{*}{Clean} & \multirow{2}{*}{FGSM} & \multirow{2}{*}{MIFGSM} & \multirow{2}{*}{\makecell[c]{PGD \\ (50-10)}} & \multirow{2}{*}{AA} & \multirow{2}{*}{\makecell[c]{C\&W \\ $l_2$}}    & \multirow{2}{*}{\makecell[c]{Time \\ (min/epoch)}} & \multirow{2}{*}{Epochs} \\
		& & & & & & & & & & \\
		\midrule[1pt]
		\multirow{2}{*}{Multi-step AT}
		& PGD-AT     & 7       & 78.0  & 50.1  & 47.4  & 43.9     & 43.4       & 71.0 & 5.4  & 100      \\
		& TRADES    & 10      & 81.0  & 55.3  & 52.4  & 49.0     & \textbf{46.4}       & 74.5 & 13.2 & 100  \\
		\cline{1-11} \TBstrut
		&FGSM-AT & 1& \textbf{88.6}&78.4 & 0.0& 0.0& 0.0& 0.0& 1.3& 80\\
		& RS-FGSM& 1&  78.0& 50.8& 48.1& 45.1& 41.5 & 73.6 & 1.3 & 30 \\
		\multirow{2}{*}{Efficient AT}
		& SubAT     & 1       & 80.6  & 50.1  & 45.7   & 43.4     & 38.1       &   74.4    &            1.3 & 30  \\
		& GradAlign & 2      & 79.5  & 52.9  & 49.9  & 46.8     & 44.1       & 74.2  &            3.3  & 30 \\
		& SSAT      & 3      & 87.8  & 50.4   & 42.7      & 33.7      & 32.2       & 41.1     &            2.1 & 30 \\
		& GAT       & 1        & 75.9  & 47.2  & 44.9  & 43.4     & 39.0       & 71.3 &             2.4 & 60  \\
		& SEAT(ours)      & 1        & 79.4 & 55.4   & 52.5  & 49.6     & 44.3       & 74.5 &           1.5 & 30 \\
		& SEAT-FL(ours) & 1 & 78.9& \textbf{56.1} & \textbf{53.8} & \textbf{50.9}& 45.8 &\textbf{75.2} & 1.5 & 30 \\
		\bottomrule[1.5pt]     
	\end{tabular}
	\label{tab:1}
\end{table*} 

\section{Experiments and Analysis}\label{sec:4}
We conduct comprehensive evaluations to verify the effectiveness of our proposed SEAT. We first present the benchmarking robustness in white-box and black-box settings, followed by extensive quantitative and qualitative analyses to provide additional insights. Furthermore, we perform all necessary evaluations to ensure that models trained using our regularization are robust and do not exhibit obfuscated gradients. Additional results are provided in \cref{Appendix:D}. 

\textbf{Training.} \,\, We follow \cite{fast} and \cite{trades} for single-step AT and multi-step AT setups. We use the same architecture, PreAct ResNet18 \cite{sat-36}, to report results across all competing defenses as well. We apply an SGD optimizer with momentum 0.9 and a weight decay of $5\times 10^{-4}$, for a total of 30 epochs.  Batch size is set 128 in all experiments. We adopt the cyclic schedule by default in single-step AT as suggested. 

\textbf{Evaluating.} \,\, Evaluation metrics we employ are \textit{Standard Accuracy} (SA) and \textit{Robust Accuracy} (RA), \textit{i.e.}, the accuracy on the original and adversarially perturbed test sets respectively. Additionally, we also employ the \textit{empirical Lipschitz constant} to measure local properties of the classifiers. We by default apply PGD-50-10 (with 50 iterations and 10 restarts) attacks with $l_\infty$ constraint for evaluation. With respect to experiments in white-box setting, we use a wide range of attacks, including $l_\infty$ and $l_2$ based attacks. Details of adversaries are provided in \cref{Appendix:D.1}. 

%		& RS-FGSM      & 1       & 80.5  & 53.7  & 49.7   & 45.6     & 42.7      & 75.8 &      79  & 30       \\

\subsection{Benchmarking the State-of-the-art Robustness}\label{sec:4.1}

\subsubsection{Performance in White-box Setting}
To close in on the true robustness, we evaluate the proposed method under a wide range of attacks in white-box setting. We compare it with several existing defenses, including multi-step AT and single-step AT methods, as shown in \cref{tab:1}. The primary dataset used for all our experiments is CIFAR-10 \cite{cifar}. In additional, we verify our scalability to larger dataset on Tiny-ImageNet \cite{tiny}. 

It can be observed that models trained using FGSM are susceptible to multi-step attacks. Unlike FGSM-AT, RS-FGSM is robust against both single-step and multi-step adversaries. However, it heavily relies on iudiciously designed learning rate schedule to get rid of catastrophic overfitting \cite{sub}. While  GAT~\cite{gat} and SSAT~\cite{co} also obtains gains in robustness when compared to FGSM-AT,they suffer from a larger accuracy-robustness trade-off than others, which might be caused by the overly severe constraint they imposed. In particular, they introduces an $l_2$ regularizer to restrict the divergence of the network outputs along all dimensions, thereby hindering the model from finding the true underlying function \cite{nuat}. Method stands out among the rest is GradAlign~\cite{gradalign}, which indeed achieves significant improvement over the prior arts. Nevertheless, it requires far more time for training when compared to other efficient AT methods.

Whereas multi-step defenses, \textit{i.e.}, PGD-AT \cite{madry} and TRADES \cite{trades} are more accurate and robust, they require much overhead, as shown in the columns of \textit{Steps}, \textit{Time} and \textit{Epochs} in  \cref{tab:1}. By contrast, our single-step method shows an almost equivalent performance to multi-step defenses, while requiring merely a fraction of the computation for training. For example, SEAT suggests a short training period with only 30 epochs in a total 50 minutes on CIFAR-10. We refer to the proposal including Flooding \cite{flood} as SEAT-FL,  which is shown to be able to lead an even larger margin over others.
\begin{figure*}[h]
	\centering
	\includegraphics[width=\textwidth]{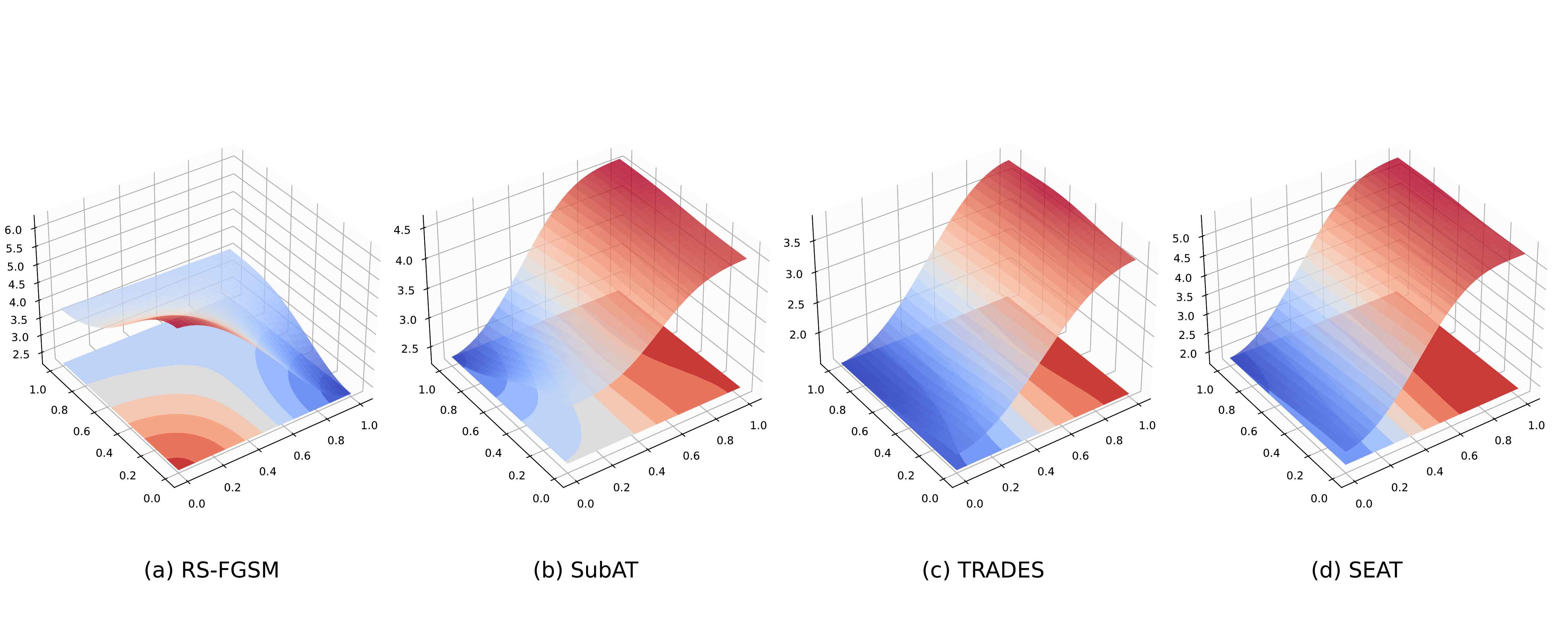}
	\caption{Input loss landscape. Visualization of loss landscape on CIFAR-10 test set for various models. We plot the softmax cross entropy loss on randomly selected input, $x^\prime = x + k \cdot g + v \cdot r$, where respectively $x$ is the origin input, $g$ is the sign of the gradient direction and $r$ (Rademacher) is the random direction, $k, v\in [0,1]$.}
	\label{Fig:inputloss}
\end{figure*}
\subsubsection{Excluding Obfuscated Gradients}
Furthermore, we present throughout evaluations on black-box attacks, gradient-free attacks and targeted attacks in \cref{Appendix:D}, to rule out obfuscated gradients as suggested by \cite{gat-3}. All of them together provide strong evidence that our defense is genuine robust.
%Furthermore, we present throughout evaluations on black-box attacks, gradient-free attacks and targeted attacks in \cref{Appendix:D}. All of them are necessary to rule out obfuscated gradients in our proposal, as suggested by \cite{gat-3}. \deleted[id=david]{Results in \ref{tab:Appendixbb}, \ref{tab:Appendixtar}, and \ref{tab:l2} all together provide strong evidence that our defense is genuine robust. }

Results in \cref{tab:2} show the performance of models in various black-box settings. It can be observed that the performance of models trained using PGD-AT, TRADES, and SEAT(ours) in black-box settings is better than that in white-box settings (as presented in  \cref{tab:1}). This confirms the absence of gradient masking. By contrast, the model trained using RS-FGSM is more vulnerable to attacks in a black-box setting than in a white-box setting. 
\begin{table}[h]\small
	\centering
	\caption{Performance (\%) of different defenses against \textbf{Black-Box} and \textbf{gradient-free attacks}. Source model for attack is speciﬁed in the column headings, and target model is speciﬁed in each row.}
	\begin{tabular}{l c c c c}
		\toprule[1.5pt]
		& \multicolumn{4}{c}{CIFAR-10}                                       \\
		\multirow{-2}{*}{Method}& Square & ResNet-18 & ResNet-34 & \multicolumn{1}{c}{VGG11} \\
		\hline
		\multicolumn{5}{c}{\cellcolor{header} Multi-step AT } \TBstrut \\
		\hline \TBstrut
		PGD-AT         & 37.9         & 34.05     & 59.98     & 75.41                       \\
		TRADES & \textbf{52.1}   & \textbf{72.84}     & 57.81     & \textbf{83.37}                     
		\\
		\hline
		\multicolumn{5}{c}{\cellcolor{header} Single-step AT } \TBstrut \\
		\hline \TBstrut
		RS-FGSM   & 0     & 31.16     & 53.53     & 67.18                         \\
		SEAT(ours)      & 51.8            & 61.87     & \textbf{62.79}     & 78.18                      \\
		\bottomrule[1.5pt]
	\end{tabular}
	\label{tab:2}
\end{table}
\subsubsection{Scalability to Larger Datasets} The scalability of adversarial training method to larger datasets has been a long-standing challenge, especially in single-step defenses. Here we verify the effectiveness of the proposed method on Tiny-ImageNet dataset \cite{tiny}. \cref{tab:tiny} in Appendix shows that SEAT outperforms state-of-the-art single-step AT and surpasses multi-step AT defense.

\subsection{Quantitative and qualitative analyses}\label{sec:4.2}

\subsubsection{Empirical study on local Lipschitz}

% As observed in \cite{look}, the local Lipschitzness shows up mostly correlated with robust accuracy; the more robust defenses tend to be the ones that impose higher degree of local Lipschitzness. So we perform experiments to verify our method, through the lens of local smoothness using the empirical Lipschitz constant. A lower value implies a smoother classifier. Details of local Lipschitzness theory are in \cref{Appendix:E}. 
% \begin{table}[h]
% 	\centering
% 	\caption{Local Lipschitz bounds of various methods on CIFAR-10 test set.}
% 	\begin{tabular}{l c c c}
% 		\toprule[1.5pt]
% 		Defense     & Lower Bound & Upper Bound & RA (\%)\\
% 		\midrule[1pt]
% 		AT          & 22.1446     & 282.1464  &  44.23\\
% 		TRADES      & \textbf{8.8810}       & \textbf{106.9340} &   \textbf{49.39} \\ \cline{1-4} \tabularnewline
% 		RS-FGSM & 414.035    & 2878.518  & 0.00\\
% 		GAT         & 21.5200       & 266.3311    & 43.54\\
% 		GradAlign   & 20.2358     & 267.3738   & 47.19\\
% 		SEAT(ours) & \textbf{19.4576}     & \textbf{254.4782} & \textbf{53.06}\\
% 		\bottomrule[1.5pt]
% 	\end{tabular}
% 	\label{tab:lp-test}
% \end{table}

% We compare the empirical Lipschitzness of classifiers trained using various defenses. Results in  \cref{tab:lp-test} demonstrate that our method produces not only the smoothest but also the robustest classifiers among them, which echoes with the observation in \cite{look} on the co-occurrence of smoothness and robustness. 
As proposed by \cite{look}, the local Lipschitzness of the underlying function can be measured by the \textit{empirical Lipschitz constant}, which can be defined as:
\begin{align}
	\dfrac{1}{n} \sum_{i=1}^{n} \max \limits_{x+\delta \in \mathcal{B}_\epsilon(x)} \dfrac{\Vert f(x_i+\delta) - f(x_i) \Vert_1}{\Vert x_i+\delta - x_i\Vert_\infty}
\end{align}
\begin{figure*}[t]
	\centering
	\setcounter{figure}{3}
	\includegraphics[width=\linewidth]{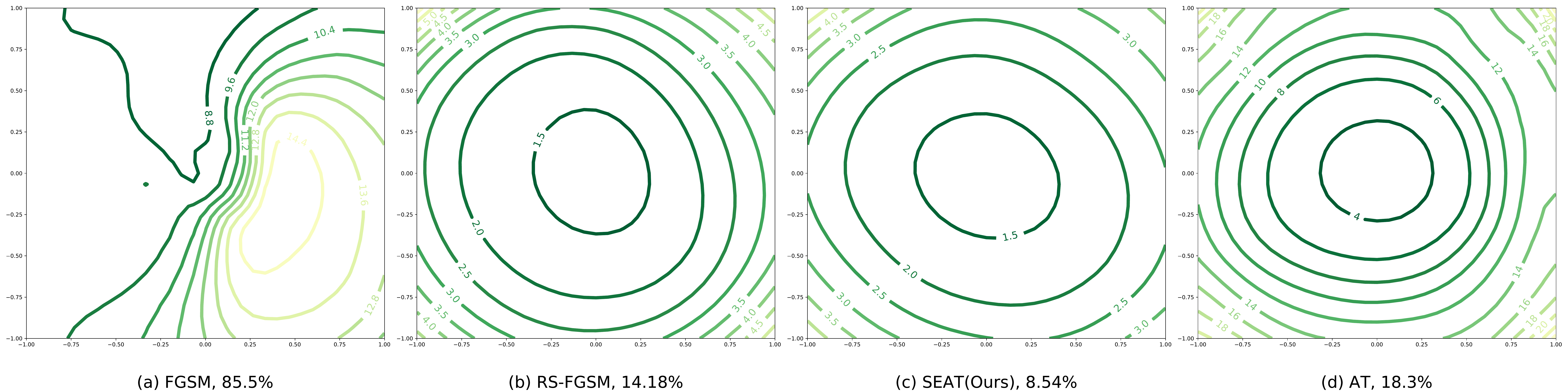}
	\caption{2-D visualization of weight loss landscape of different methods on the CIFAR-10 test set. We also present the gap of robust generalization, \textit{i.e.}, the difference between the training and test robust accuracy. A lower value indicates a better generalization.}
	\label{Fig:2d_main}
\end{figure*}

\noindent We can identify a smoother classifier by a lower value of this constant. Here, we respectively calculate the lower bound and upper bound in \cref{tab:3} by minimizing and maximizing the formula above. To delve into our regularizer and provide further insights, we vary lambda $\lambda$ that serving as a controller on the strength of linearization, from 0 to 0.7. Such evaluation further verifies that our regularizer indeed imposes local Lipschitzness, with smoothness improved with a higher  lambda $\lambda$ overall. But we also note that a still higher lambda $\lambda$ appears to hurt the performance. Upon these experimental results, we set $\lambda=0.5$, and also for a well balance.
\begin{table}[t]\small
	\centering
	\caption{Local Lipschitz bound and generalization gap of the proposed SEAT with varying $\lambda$ in \cref{eq:sec3.8}.}
	\begin{tabular}{l c c c c c}
		\toprule[1.5pt]
		$\lambda$ & Lower Bound & Upper Bound & SA &  RA & Gap \\
		\midrule[1pt]
		0      & 22.31     & 286.64    & \textbf{88.41}     & 46.29   & 42.12              \\
		0.1    & 23.05     & 293.50    & 86.68     & 48.31   & 38.37              \\
		0.3    & 21.26     & 268.51    & 82.16     & 50.27   & 31.88              \\
		0.5    & \textbf{20.33}     & \textbf{260.11}    & 81.12     & \textbf{52.47}   & \textbf{28.74}              \\
		0.7    & 21.72      & 277.85    & 81.22     & 51.55   & 29.67              \\
		\bottomrule[1.5pt]
	\end{tabular}
	\label{tab:3}
\end{table}

\subsubsection{Impoved local properties}
To verify whether our regularizer improves local properties, we visualize the loss landscape in the vicinity of a randomly chosen test input. To obtain the plots, we perturb the input along two directions: one along the adversarial perturbation, and the other along a random direction.  \cref{Fig:inputloss} shows that the surface of single-step defense RS-FGSM is undersirably curved, implying sensitiveness to perturbations. Another state-of-the-art SubAT \cite{sub} also exhibits convolved surface. In contrast, models trained using SEAT and TRADES produce smooth and flat landscape, thereby satisfying robustness. 

Note that TRADES is an iterative method, whose desirable surface is achieved by adding a perturbation to the input at every step for a monotonically increasing loss, and thus to mitigate the overfitting. Remarkably, our single-step training method requires much lower overhead while produces surface akin to that of the expensive multi-step defenses, thereby achieving impressing gains in robustness.% We would further expound this in the Supplementary section. 

\subsubsection{Enhanced Robustness with Flatter Minima}
\begin{figure}[h]
	\centering
	\setcounter{figure}{2}
	\includegraphics[width=\linewidth]{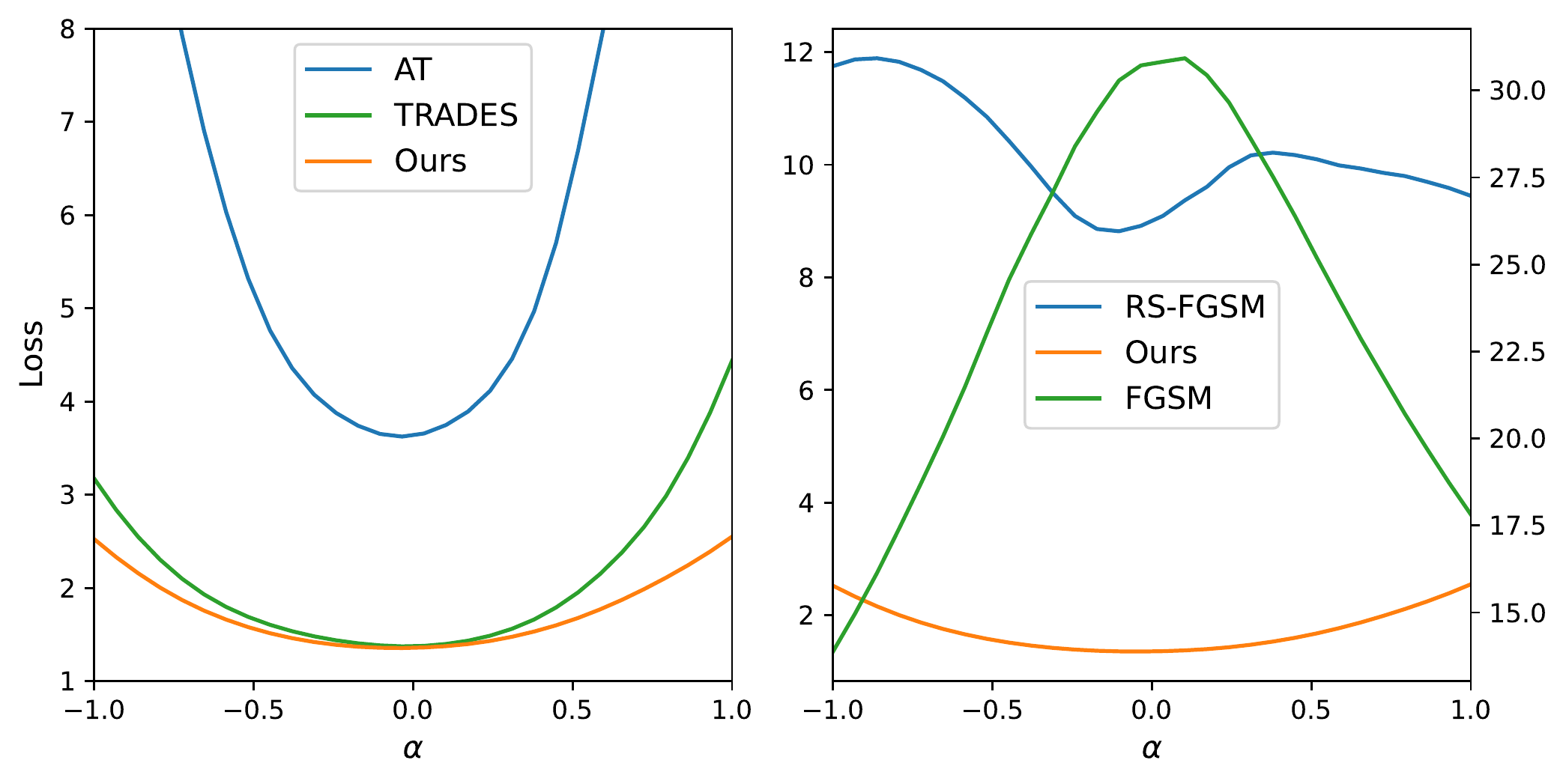}
	\caption{Weight loss landscape obtained from the PreAct ResNet18 \cite{amp-18} models trained with different defenses on CIFAR-10 test set.}
	\label{Fig:weightloss}
\end{figure}
To delve into our method, we go beyond the input loss landscape to analytically investigate the weight loss landscape, which depicts the geometry of the loss landscape around model weights instead of the randomly sampled input. It is useful for indicating the robust generalization, \textit{i.e.}, the difference between training and test robustness, as a flatter weight loss landscape causes the model to generalize better \cite{sub-40-awp}. As shown in  \cref{Fig:weightloss}, we found the proposed SEAT performs the best no matter being compared with multi-step AT (left) or single-step AT(right). The flattest local minima indicating the least sensitive to weight perturbations, and thus the best robust generalization. Notably, the loss of FGSM and RS-FGSM present drastic changes, revealing the occurrence of catastrophic overfitting.

We further visualize the 2-D weight loss landscape in  \cref{Fig:2d_main}, which shows that SEAT produces the flatter weight loss landscape around the local optima than other methods. We reach the same conclusion through both 1-D visualization and 2-D visualization. It means that our method could reliably find the flat minima and therefore achieve better robust generalization. 
\subsubsection{Absence of Decision Boundary Distortion} As demonstrated by \cite{co}, catastrophic overfitting leads to \textit{decision boundary distortion}. This is a phenomenon in which a model appears to be robust against adversarial example with the maximum perturbation, yet loses its robustness to a smaller magnitude. As shown in \cref{Fig:main_decision}, model trained using FGSM-AT reveals distorted interval, implying that it is overfitted for FGSM attack but vulnerable to multi-step adversaries that can search more areas within the budget. Unlike FGSM-AT, the proposed SEAT is able to escape from decision boundary distortion, with significantly flatten loss surface and more areas covering correct decision. 
\begin{figure}[t]
	\centering
	\setcounter{figure}{4}
	\includegraphics[width = \linewidth]{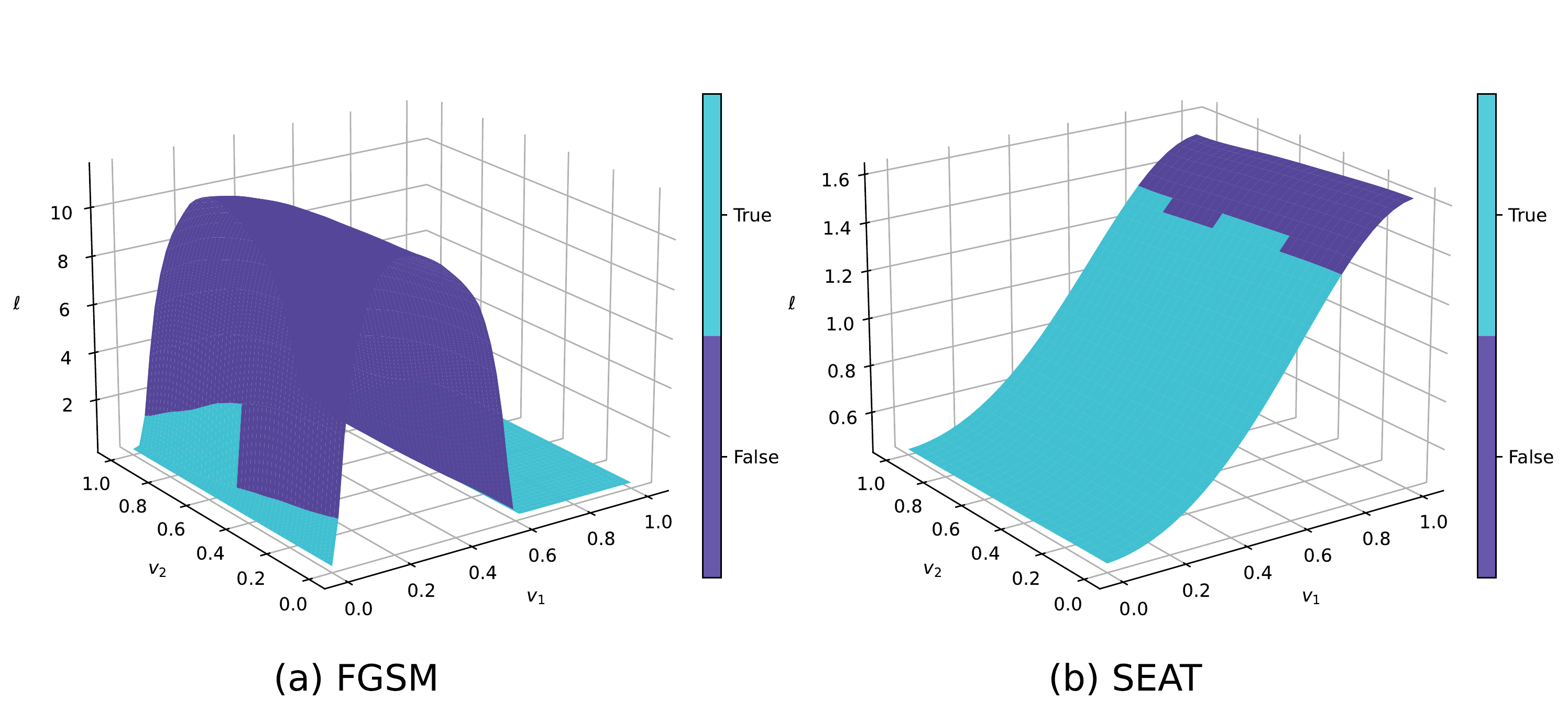}
	\caption{Visualization of decision boundary. A highly distorted boundary can be observed in FGSM trained model.}
	\label{Fig:main_decision}
\end{figure}
\subsubsection{Reliability of SEAT}
\begin{figure}[t]
	\centering
	\includegraphics[width=\linewidth]{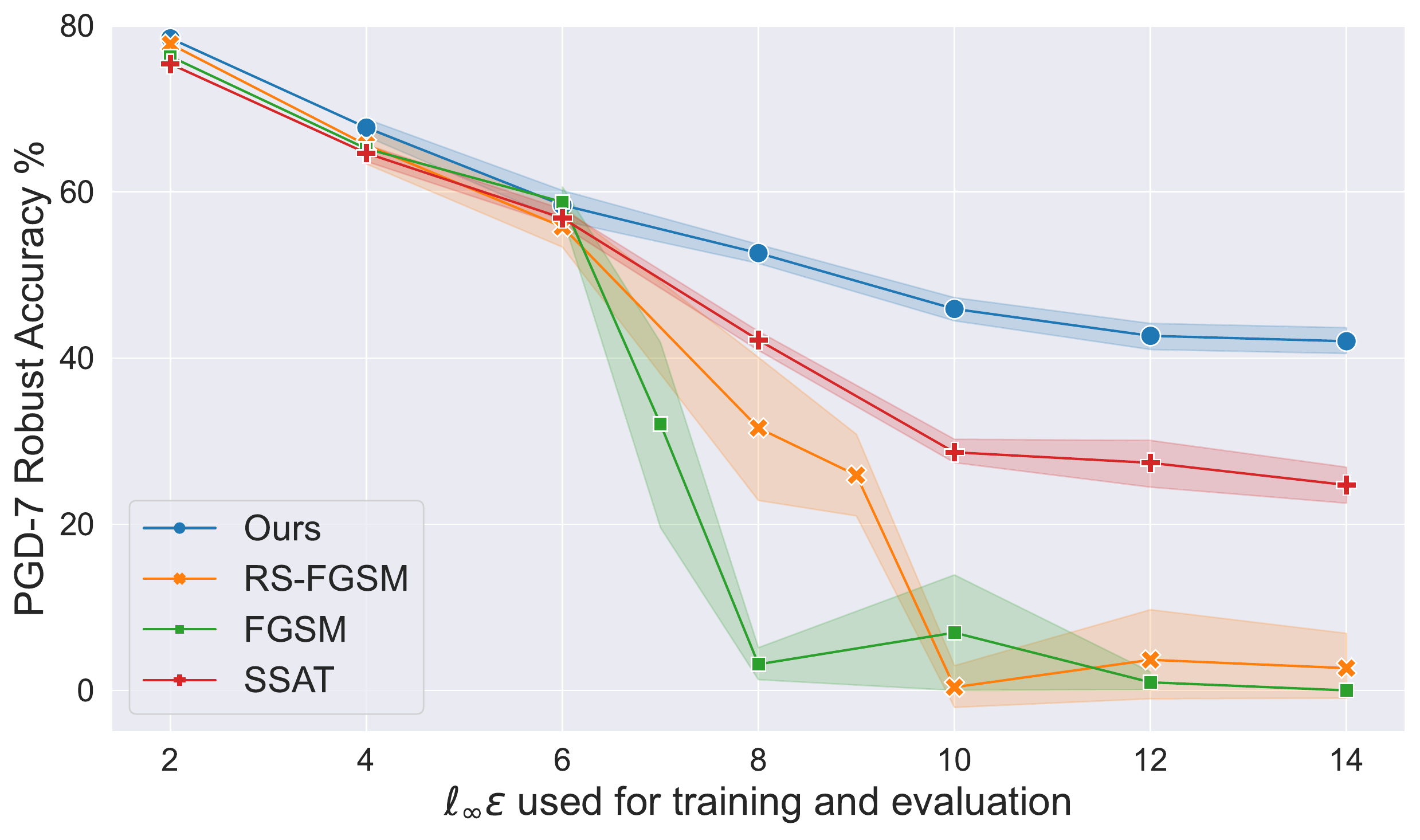}
	\caption{Robust Accuracy of various methods on CIFAR-10 with Pre-Act ResNet-18. The results are averaged over 5 different random seeds.}
	\label{Fig:reliably}
\end{figure}

Multi-step AT using a larger radius renders better robustness to some extend. Now, we would like to investigate whether larger radius could help in single-step AT. As  \cref{Fig:reliably} shown, neither FGSM nor RS-FGSM benefit from larger training radius $\epsilon$ due to catastrophic overfitting. While SSAT \cite{co} is able to take advantages of larger radius, it obtains inferior robustness compared to our method. Our result reveals that SEAT successfully avoids catastrophic overfitting, and also allows single-step adversarial training with larger radius.

\subsection{Ablation studies}\label{sec:4.3}
As the proposed method involves modiﬁcations on both the maximization and outer minimization, in what follows, we perform ablations study with respective to the attack generation and adversarial defense.
\begin{table}[h]\small
	\centering
	\caption{SEAT ablations. We report the robust performance (\%) of ablations of SEAT under various attacks, which covering both $l_\infty$ and $l_2$ threat models.}
	\begin{tabular}{l c c c c}
		\toprule[1.5pt]
		Ablations                                     &FGSM& PGD   & AA & C\&W  \\
		\midrule[1pt]
		\textbf{Full model}                                          & \textbf{57.23} & \textbf{54.08} & \textbf{45.8} & 75.30\\
		\textbf{A1}: w/o proposed attack   & 54.78 &50.20  & 44.3 & 74.00\\
		\textbf{A2}: w/o linearization                   & 52.58 &46.06 & 42.2 & \textbf{77.20}\\
		\textbf{A3}: w/o Flooding                            & 55.38 &50.66 & 44.3 & 74.20\\
		\bottomrule[1.5pt]
	\end{tabular}
	\label{tab:ab}
\end{table} 
\subsubsection{Analysis on Adversarial defense}
In this part, we compare against the ablations of our full model to explore if each component is essential. Results reported in  \cref{tab:ab} demonstrate that each component of the proposed training object is indispensable for our eventual performance, as in general the robust accuracy of ablation is lower than that of our full model. Remarkably, performance get significantly deteriorated when the linearization regularizer is absent (\textbf{A2}). This suggests that our method greatly beneﬁts from function smoothing, which leads to enhanced robustness.

\subsubsection{Analysis on Attack Generation}
In addition to the ablation study above, we further evaluate SEAT by analyzing the effect of the proposed attack on loss landscape \textit{w.r.t.} input. Qualitative results are presented in \cref{Appendix:F}. We note that the landscape gets flatter with the step-by-step integration of component. This implies that both the maximum margin loss and linear regularization are indeed effective for the generation of useful attacks for training. Moreover, the corresponding contours projected from loss surface onto the horizontal plane also provide consistent observation, with contours getting less convolved, revealing improved robustness against perturbations in which along the gradient direction.

As a whole, the proposed SEAT offers impressing adversarial robustness against a broad range of adversaries and across datasets, while it requires only a lightweight training. This makes our method more appealing in practice.
\section{Conclusion}\label{sec:5}
In this work, we focus on catastrophic overfitting, which is a dominant phenomenon in single-step adversarial training of deep models. We first provide an empirical study to better understand catastrophic overfitting. Building on it, we devise our regularizer, which harnesses on the local linearity of models trained using multi-step methods, and incorporates such salient property into models trained using single-step adversarial training.  The impressing performance of SEAT arguably makes it into the current state of the art amongst single-step adversarial training methods. While multi-step adversarial training remains one of the most effective way to secure a deep network against attacks, few can afford it due to its high computation cost. Thus, our study put adversarial training within reach for organizations with modest computation budgets.

%%%%%%%%% REFERENCES
{\small
\bibliographystyle{ieee_fullname}
\bibliography{ref}
}

\clearpage
\onecolumn
\appendix

\setcounter{property}{0}
\setcounter{remark}{0}
\counterwithin{figure}{section}
\counterwithin{table}{section}
\counterwithin{equation}{section}
\counterwithin{equation}{section}

\section{Proof of the property 1}\label{Appendix:A}

\begin{property}
	Consider a loss function $\ell(\cdot)$ that is once-differentiable, and a local neighbourhood $\mathcal{B}_\epsilon(x)$ that of radius $\epsilon$ and centered at $x$. Then for any $x+\delta \in \mathcal{B}_\epsilon(x)$, we have
	\begin{align}
		\begin{split}
			\vert \ell({x+\delta}) - \ell(x)-\langle\delta,\nabla_x\ell(x)\rangle \vert \leq (\epsilon + \frac{1}{\sqrt{n}} \cdot\Vert \delta \Vert_2) \cdot \Vert \nabla\ell(x) \Vert_1.
		\end{split}
	\end{align}
\end{property}
\begin{lemma}
	For finite-dimensional sequence $x=(x_1,x_2,\cdots x_n)$ $\in \mathbb{R}^n$, the $\ell_p$ norm denotes $\Vert x \Vert_p = \big( \sum \limits_i \vert x_i \vert ^p \big)^{1/p}$. For any $0<p<q$, the $\ell_p$ and $\ell_q$ norm are equivalent on $\mathbb{R}^n$. It means:
	\begin{align}
		\Vert x \Vert_p \leq n^{1/p-1/q} \cdot \Vert x \Vert _q
	\end{align}
	\label{lemma}
\end{lemma}
\noindent \textbf{\textit{Proof.}} \,\, We would start from the local Taylor expansion \cref{eq:sec3.2}. It is clear that when $x\to x+\delta$, the higher order term $\omega(\delta)$ becomes negligible. Mostly, we instead use its equivalent, \textit{i.e.}, $\vert \ell({x+\delta}) - \ell(x)-\langle\delta,\nabla_x\ell(x)\rangle \vert$ as a measure of how linear the surface is within a neighbourhood. We call this measure the \textit{linearity approximation error} $\xi(\theta,x)$. Apparently, this is precisely the regularizer we defined in \cref{eq:sec3.5} in the main text.

\noindent Applying the \textit{Cauchy Schwarz inequality}, we have
\begin{align}
	\begin{split}
		\vert \ell({x+\delta}) - \ell(x)-\langle\delta,\nabla_x\ell(x)\rangle \vert &\leq \vert  \ell({x+\delta}) - \ell(x) 
		\vert + \vert  \langle\delta,\nabla_x\ell(x)\rangle  \vert \\
		&=\epsilon\cdot \Vert \nabla\ell(x) \Vert_1  + \vert  \langle\delta,\nabla_x\ell(x)\rangle  \vert \\
		&\leq \epsilon\cdot \Vert \nabla\ell(x) \Vert_1  + \Vert \delta\Vert_2 \cdot \Vert \nabla_x\ell(x)\Vert_2 \\
	\end{split}
	\label{eq:A3}
\end{align}
where we would like to simplify the second term.

\noindent By assigning $p=2$ and $q=1$ in \textit{Lemma} \cref{lemma}, we have
\begin{align}
	\Vert x \Vert_2 \leq \dfrac{1}{\sqrt{n}} \cdot \Vert x \Vert_1
	\label{eq:12}
\end{align}
Back to \cref{eq:A3}, the inequality could be further simplified based on \cref{eq:12}. Now we arrive at
\begin{align}
	\begin{split}
		\vert \ell({x+\delta}) - \ell(x)-\langle\delta,\nabla_x\ell(x)\rangle \vert 
		\leq (\epsilon + \frac{1}{\sqrt{n}} \cdot \Vert \delta \Vert_2) \cdot \Vert \nabla\ell(x) \Vert_1 
	\end{split}
	\label{eq:A5}
\end{align}

\begin{remark}
	From \cref{eq:A5} it is clear that by bounding the linearity approximation error, one can implicitly introduces an adaptive scaling parameter $\Vert \delta \Vert_2 / \sqrt{n}$ for perturbation generation instead of a fixed magnitude $\epsilon \cdot \Vert \nabla_x \ell(x) \Vert_1$, which is the main cause of catastrophic overfitting.
\end{remark}

\section{Improved local properties achieved by the regularizer}\label{Appendix:B}

In this section, we present details on the improved local properties induced by the regularizer. We would start from the definition of local Lipschitz continuity.

\begin{theorem}
	Given two metric space $(\mathcal{X}, d_\mathcal{X})$ and $(\mathcal{Y},d_\mathcal{Y})$, where $d_\mathcal{X}$ denotes the metric on the set $\mathcal{X}$, a function $f: \mathcal{X} \to \mathcal{Y}$ is called Lipschitz continuous if there exists a real constant $K\geq0$ such that, for all $x_1,x_2 \in \mathcal{X}$,
	\begin{align}
		d_\mathcal{Y}\big( f(x_1),f(x_2) \big) \leq K \cdot d_\mathcal{X}(x_1, x_2).
	\end{align}
	\label{metric}
\end{theorem}

\noindent Any satisfied $K$ can be referred to as a Lipschitz constant of the function $f$. Based on this framework, the Lipschitz continuous on the adversarial loss function is as follows:
\begin{align}
	\vert \ell(x+\delta)-\ell(x)\vert \leq K\cdot \Vert x+\delta - x\Vert_2 = K \cdot 
	\Vert \delta \Vert _2,
\end{align}
where $\ell(\cdot)$ denotes the loss function $\ell:\mathcal{X}\to \mathbb{R}$, $x\in \mathcal{X}$ represents input and $x+\delta$ is the adversarial example. Since we target for robustness against adversaries that are lying within a $\ell_p$-ball of radius $\epsilon$ and centered at $x$, we would expect a loss function that is local Lipschitz continuous within: 
\begin{align}
	\mathcal{B}_\epsilon(x) = \{ x+\delta: \Vert\delta \Vert_\infty \leq \epsilon \}
\end{align}
Given any two points in $\mathcal{B}_\epsilon(x)$, the local Lipschitz constant $K$ of interest is given by:
\begin{align}
	K = \sup \limits_{x+\delta \in \mathcal{B}_\epsilon(x)} \dfrac{\vert \ell(x+\delta) - \ell(x) \vert }{\Vert \delta  \Vert_2}
\end{align}
Applying a subtraction on both sides, we have,
\begin{align}
	\begin{split}
		K - \dfrac{\vert \langle \delta,\nabla_x\ell(x) \rangle \vert }{\Vert \delta \Vert_2}  &=
		\sup \limits_{x+\delta \in \mathcal{B}_\epsilon(x)} \dfrac{\vert \ell(x+\delta) - \ell(x) \vert }{\Vert \delta \Vert_2} - \dfrac{\vert \langle \delta,\nabla_x\ell(x) \rangle \vert}{\Vert \delta \Vert_2} \\
		&< \sup \limits_{x+\delta \in \mathcal{B}_\epsilon(x)} \dfrac{\vert \ell(x+\delta) - \ell(x)-\langle \delta,\nabla_x\ell(x)\rangle \vert }{\Vert \delta \Vert _2}, \\
		K &< \sup \limits_{x+\delta \in \mathcal{B}_\epsilon(x)} \dfrac{\vert \ell(x+\delta) - \ell(x)-\langle \delta,\nabla_x\ell(x)\rangle \vert + \vert \langle \delta,\nabla_x\ell(x) \rangle \vert }{\Vert \delta \Vert _2}
	\end{split}
\end{align}
The \textit{linearity approximation error} $\xi(\theta, x)$ composes the expression on the RHS. Hence, minimizing $\xi(\theta, x)$ is expected to induce a smaller locally Lipschitz constant \textit{K}, thereby encouraging the optimization procedure to find a model that is locally Lipschitz continuous. 

\section{Empirical study}\label{Appendix:C}

\subsection{Empirical observations on the decision boundary}\label{Appendix:C.1}
\begin{figure}[h]
	\centering
	\includegraphics[width=0.75\textwidth]{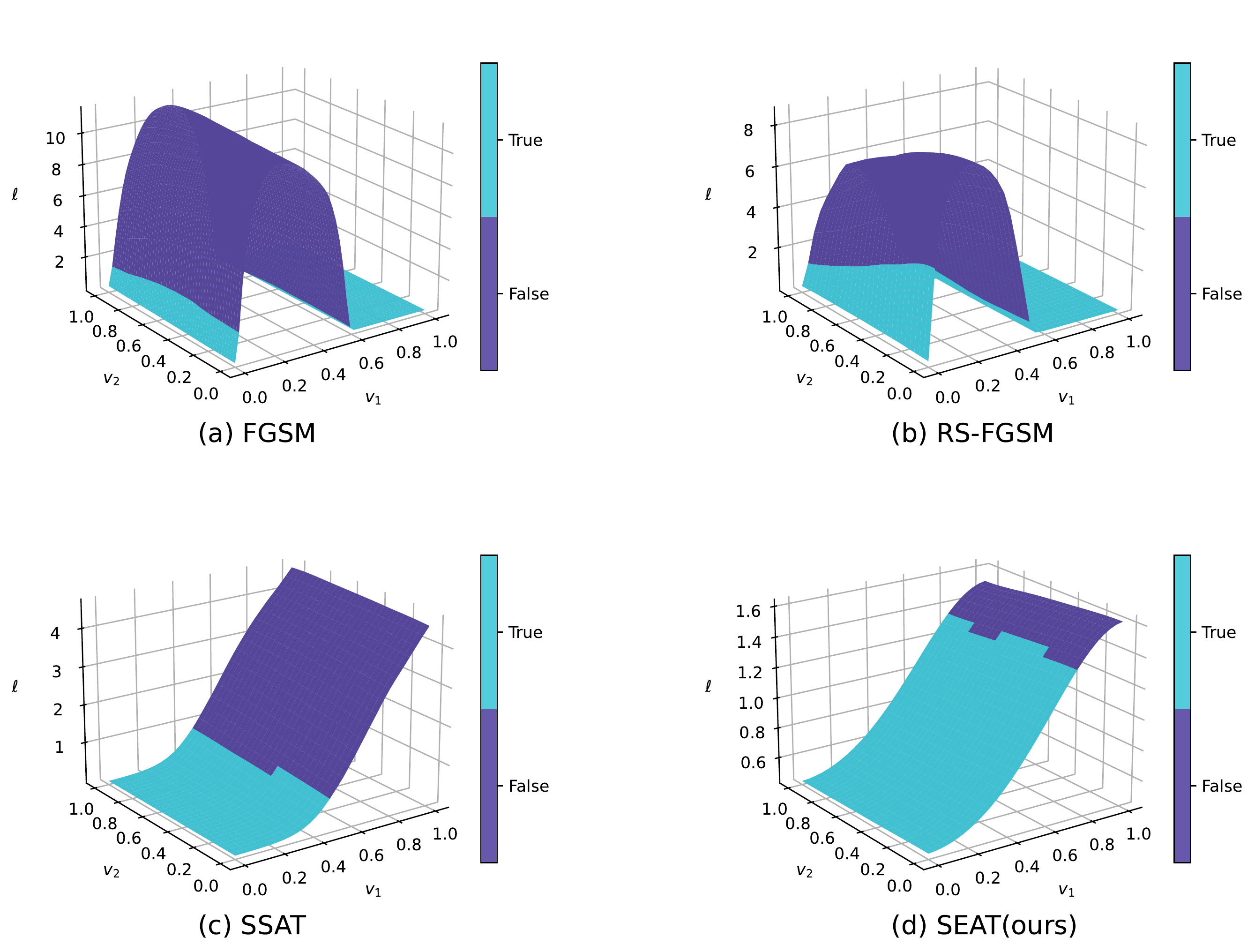}
	\caption{Visualization of decision boundary. The $r_1$ axis is the direction of the FGSM perturbation and $r_2$ is a random direction. We display the cross entropy loss of the adversarial input $x^\prime = x + k \cdot r_1 + v \cdot r_2$, where $k,v \in [0, 1]$.}
	\label{Fig:decision}
\end{figure}

In this part, we present more figures to demonstrate that our proposed method can successfully avoid decision boundary distortion. As described in \cref{sec:4.2} of our main text, distorted intervals can be used to indicate whether catastrophic overfitting occurs. As shown in  \cref{Fig:decision}, both FGSM and RS-FGSM are catastrophic overfitted models, as the intervals indicated. On the contrary, such distortion is absent in both SEAT and SSAT, similar to the prior art SSAT \cite{co}. Notably, our model achieves the most desirable decision boundary than other competing methods, with nearly flat surface and also large areas of correct decision.

%\begin{figure}[t]
%	\centering
%	\subfloat[FGSM]{\includegraphics[width = 0.45\textwidth]{fig/fgsm_decision.eps}}
%	\subfloat[FastAT]{\includegraphics[width = 0.45\textwidth]{fig/fast_decision.eps}}
%	\newline
%	\subfloat[FGSM-CKPT]{\includegraphics[width = 0.45\textwidth]{fig/understanding_decision.eps}}
%	\subfloat[Ours]{\includegraphics[width = 0.45\textwidth]{fig/ours_decision.eps}}
%	\label{fig:6}
%	\caption{The $r_1$ axis is the direction of the FGSM perturbation and $r_2$ is a random direction. We compute the cross entropy loss of the adversarial input $x^\prime = x + k \cdot r_1 + v \cdot r_2$, where $k,v \in [0, 1]$.}
%\end{figure}

\subsection{Empirical observations on local linearity of the loss function}\label{Appendix:C.2}

 \cref{Fig:inputloss} in the main text shows the highly curved loss landscape of model trained with single-step attacks, implying the occurrence of gradient masking. As we mentioned in \cref{sec:3.1}, the generation of single-step attacks such as FGSM is based on the liner approximation of the loss function. Nevertheless, the gradient masking causes the linear assumption become unreliable, which further induces catastrophic overfitting. 

To detect catastrophic overfitting from the perspective of local linearity reduction, we propose to measure the non-linearity of the loss function. To this end, we use a local linearity error, which is derived form the first-order Taylor expansion and can be calculated as:
\begin{align}
	\left | \ell\left ( x+\delta  \right ) -\ell\left ( x \right )  -\delta ^{T}\nabla _{x} \ell\left ( x \right ) \right |
\end{align}

\begin{figure}[h]
	\centering
	\includegraphics[width=0.6\textwidth]{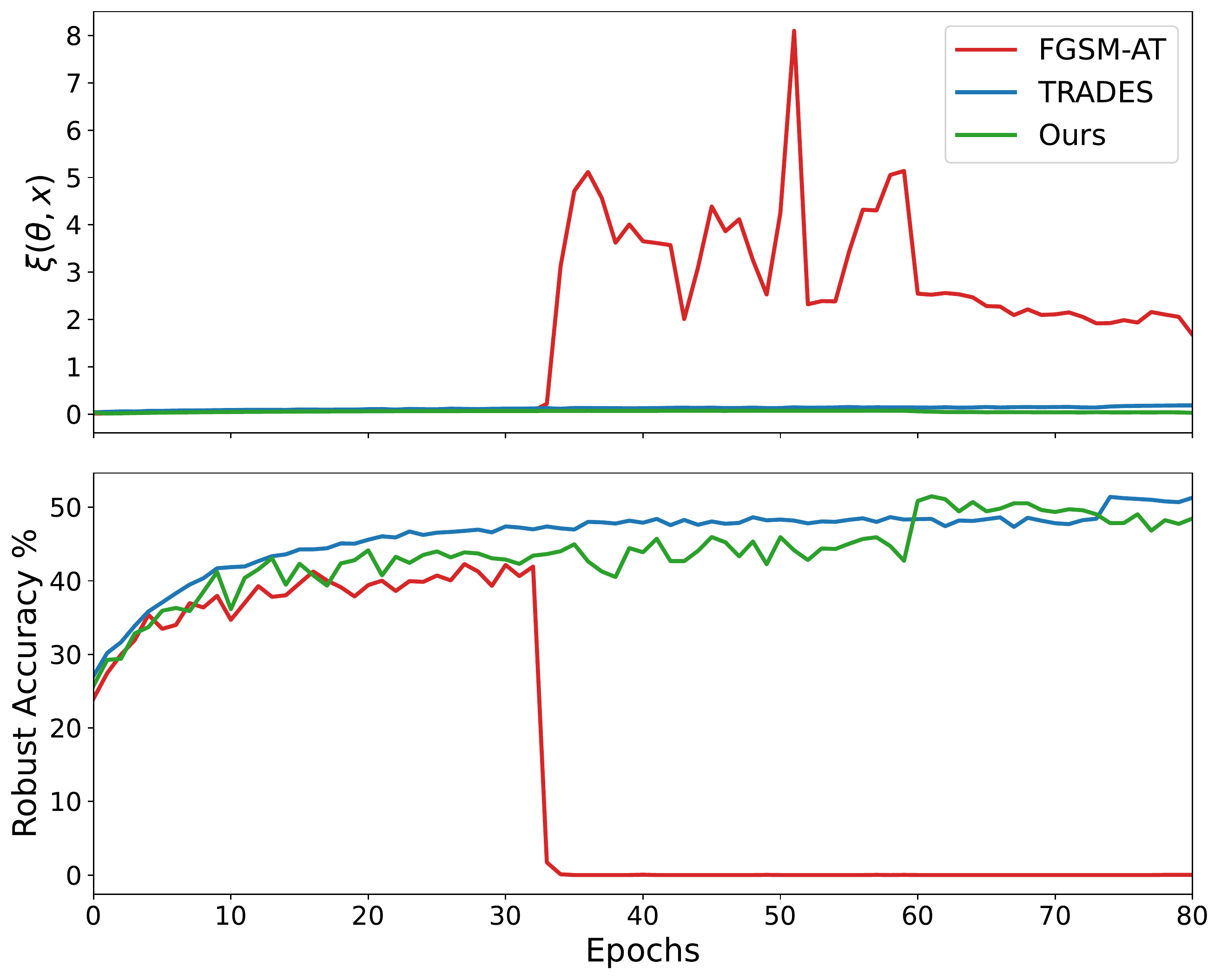}
	\caption{The local linearity of the loss function. Top: Robust accuracy against PGD-20 on the training images during the training process. Bottom: The average of the local linearity error over the entire training set.}
	\label{Fig:appendix}
\end{figure}
 \cref{Fig:appendix} shows that catastrophic overfitting coincides with the local linearity reduction of the loss function. To be more specific, the linearity approximation error $\xi(\theta,x)$ of FGSM-AT abruptly increases in the moment when the catastrophic overfitting occurs, \textit{i.e.}, the model suddenly loses its robustness against PGD attacks during training within an epoch and eventually collapses.

We further note that, the proposed single-step adversarial training method SEAT behaves in a strikingly similar way to that of TRADES, which generates adversarial examples with multiple iterations and thus to avoid catastrophic overfitting. Both our SEAT and TRADES bring in negligible values of $\xi(\theta,x)$ during the training, indicating our impressing maintenance of the local linearity and thus to mitigate catastrophic overfitting.  In contrast, the fast adversarial training baseline FGSM-AT reports noticeable large values. This implies the linearity assumption is violated-the loss surface of the trained model is highly bumpy, which obfuscates the direction for strong attack generation.   

\section{Additional experimental results}\label{Appendix:D}

\subsection{Details on Attack Methods}\label{Appendix:D.1}
We verify the proposed method over a broad range of adversaries, including FGSM \cite{sat-7}, MI-FGSM \cite{sat-4}, PGD \cite{madry}, Carlini \& Wagner \cite{sat-15-cw} and DeepFool \cite{sat-13}. We also consider Auto-Attack \cite{aa}, which consists of APGD-CE, APGD-DLR, FAB \cite{fab} and Square \cite{square}. 

With respective to perturbation size, we set step size $\alpha=1.25\epsilon$ for single-step adversarial training. As for multi-step training, $\alpha=2/255$ and $\alpha=15/255$ are respectively used for the generation of  $l_\infty$ attacks and $l_2$ attacks, as suggested by \cite{fast}.

\subsection{Performance against targeted attacks}\label{Appendix:D.3}
We verify our method on both 10-step and 20-step PGD targeted attacks. Here we consider Random target attack and the Least Likely attack for evaluation. We also present the results in untargeted setting for comparison, as shown in  \cref{tab:Appendixtar}. As untargeted attack is more challenging than targeted attack for models to defend against, the robust accuracy against the targeted attack is correspondingly higher than the baseline set by untargeted attacks.
\begin{table}[h]
	\centering
	\caption{Performance (\%) of the proposed SEAT in various targeted White-Box attack settings on CIFAR-10 dataset. We additionally provide the performance in untargeted setting for comparison.}
	\begin{tabular}{l c c}
		\toprule[1.5pt]
		\multirow{2}{*}{Attack}  & \multicolumn{2}{c}{CIFAR-10} \\
		& 10-steps      & 20-steps     \\ 
		\midrule[1pt]
		PGD-Target(Least Likely) & 73.08         & 73.05        \\
		PGD-Target(Random)       & 70.24         & 70.11        \\
		PGD-Untarget             & 53.98         & 53.02        \\
		\bottomrule[1.5pt]
	\end{tabular}
	\label{tab:Appendixtar}
\end{table}

\subsection{Performance against $l_2$ attacks}\label{Appendix:D.4}
We also evaluate our method against $l_2$ attacks, and report the robust performance on two metrics, \textit{i.e.}, fooling rate and mean $l_2$ norm. We use DeepFool \cite{sat-13} and  C\&W attack \cite{sat-15-cw}, which are generated without norm constraints. Generally speaking, it requires a larger $l_2$ norm to fool a robust classifier than an undefended model. As for the fooling rate, it calculates the percentage of perturbed testing samples that are misclassified, thus can be used for indicating robustness. As shown in  \cref{tab:l2}, model trained using the proposed method consistently improves over the single-step baseline method. 
\begin{table}[h]
	\centering
	\caption{Performance (\%) of different defenses against DeepFool and C\&W attacks. For the mean $l_2$ norm, higher the better. With respect to fooling rate, as it defines the percentage of testing samples that have been misclassified, the lower the better.}
	\begin{tabular}{l c c c c}
		\toprule[1.5pt]
		\multirow{2}{*}{Method} & \multicolumn{2}{c}{DeepFool} & \multicolumn{2}{c}{CW} \\
		& Fooling Rate       & Mean $l_2$      & Fooling Rate    & Mean $l_2$   \\ \midrule[1pt]
		RS-FGSM          \cite{fast}       & 99.5          & 0.97         & 100        & 0.2       \\
		SEAT(ours)                    & 99.5          & 0.99            & 46         & 0.3   
		\\
		PGD-7  \cite{madry}                 & 99.0          & 0.92         & 72         & 0.5       \\
		\bottomrule[1.5pt]  
	\end{tabular}
	\label{tab:l2}
\end{table}

\subsection{Scalability to Tiny-ImageNet}\label{Appendix:D.5}
Performance against white-box attacks for Tiny-ImageNet are shown in  \cref{tab:tiny}. We also similarly observe superior performance amongst singe-step defenses and comparable robustness with the multi-step baseline. Notably, SEAT is able to strike a more favourable trade-off between accuracy and robustness.   
\begin{table}[h]
	\centering
	\caption{White-box evaluation (Tiny-ImageNet). Accuracy (\%) of different defense (rows) trained on PreAct ResNet18 against $l_\infty$ norm bound ($\epsilon=8/255$) white-box attacks.}
	\begin{tabular}{c l c c c}
		\toprule[1.5pt]
		&Methods     & Steps & SA    & PGD-20 \\
		\midrule[1pt]
		Multi-step & PGD-AT \cite{madry}    & 10   & 42.76 & 19.84     \\ \cline{1-5}
		\TBstrut		
		\multirow{3}{*}{Single-step} &RS-FGSM  \cite{fast}   & 1    & 28.79 & 11.90      \\
		&Sub-AT \cite{sub} & 1    & 39.38 & 16.75     \\
		&SEAT(ours)        & 1    & 49.19 & 19.39     \\
		\bottomrule[1.5pt]
	\end{tabular}
	\label{tab:tiny}
\end{table}

%\clearpage
\section{Measuring Local Lipschitzness}\label{Appendix:E}
As observed in \cite{look}, the local Lipschitzness shows up mostly correlated with robust accuracy; the more robust defenses tend to be the ones that impose higher degree of local Lipschitzness. So we perform experiments to verify our method, through the lens of local smoothness using the empirical Lipschitz constant. A lower value implies a smoother classifier.
\begin{table}[h]
	\centering
	\caption{Local Lipschitz bounds of various methods on CIFAR-10 test set.}
	\begin{tabular}{l c c c}
		\toprule[1.5pt]
		Defense     & Lower Bound & Upper Bound & RA (\%)\\
		\midrule[1pt]
		AT          & 22.1446     & 282.1464  &  44.23\\
		TRADES      & \textbf{8.8810}       & \textbf{106.9340} &   \textbf{49.39} \\ \cline{1-4} 
		\TBstrut
		RS-FGSM & 414.035    & 2878.518  & 0.00\\
		GAT         & 21.5200       & 266.3311    & 43.54\\
		GradAlign   & 20.2358     & 267.3738   & 47.19\\
		SEAT(ours) & \textbf{19.4576}     & \textbf{254.4782} & \textbf{53.06}\\
		\bottomrule[1.5pt]
	\end{tabular}
	\label{tab:lp-test}
\end{table}

We compare the empirical Lipschitzness of classifiers trained using various defenses. Results in \cref{tab:lp-test} demonstrate that our method produces not only the smoothest but also the robustest classifiers among them, which is echoes with the observation in \cite{look} on the co-occurrence of smoothness and robustness. 

\section{Ablation studies}\label{Appendix:F}
As the proposed method involves modiﬁcations on both the inner maximization and outer minimization, in what follows, we perform ablations study with respective to the attack generation and adversarial defense, to explore the effect of each component in our proposal.
\subsection{Analysis on Attack Generation}
\begin{figure}[h]
	\subfloat[CE]{\includegraphics[width = 0.32\textwidth]{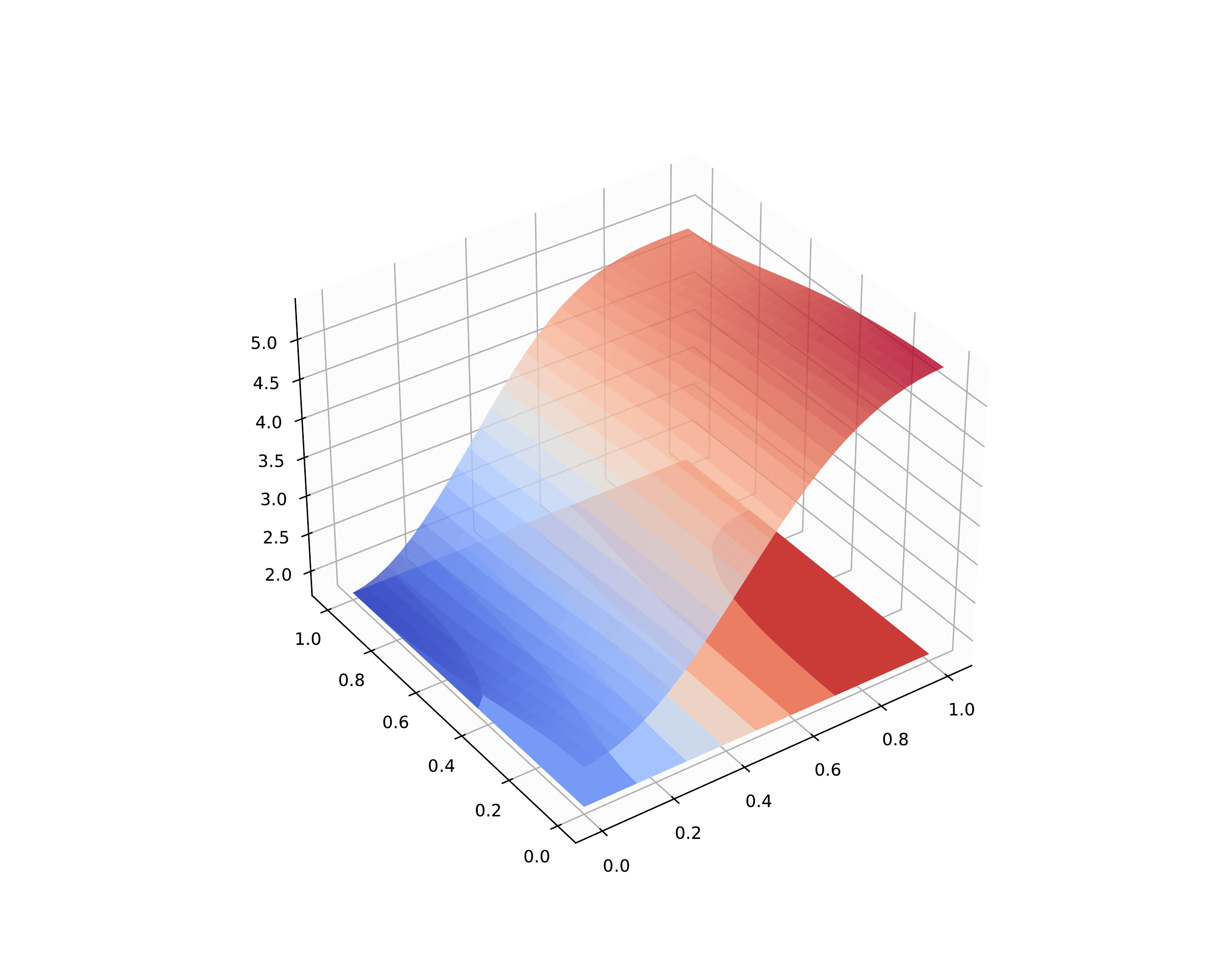}}
	\subfloat[CE + Linearization]{\includegraphics[width = 0.32\textwidth]{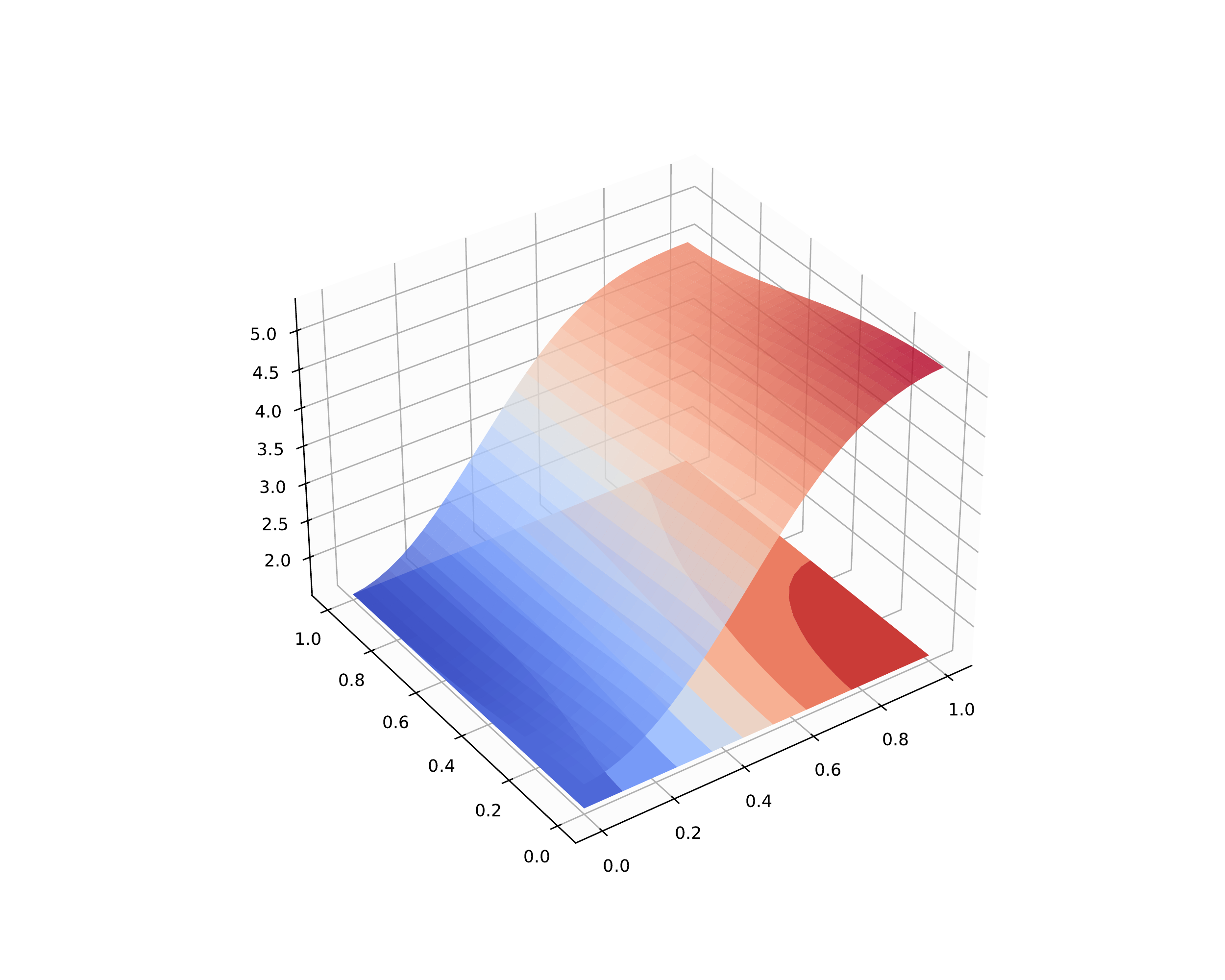}}
	\subfloat[Margin loss + Linearization]{\includegraphics[width = 0.32\textwidth]{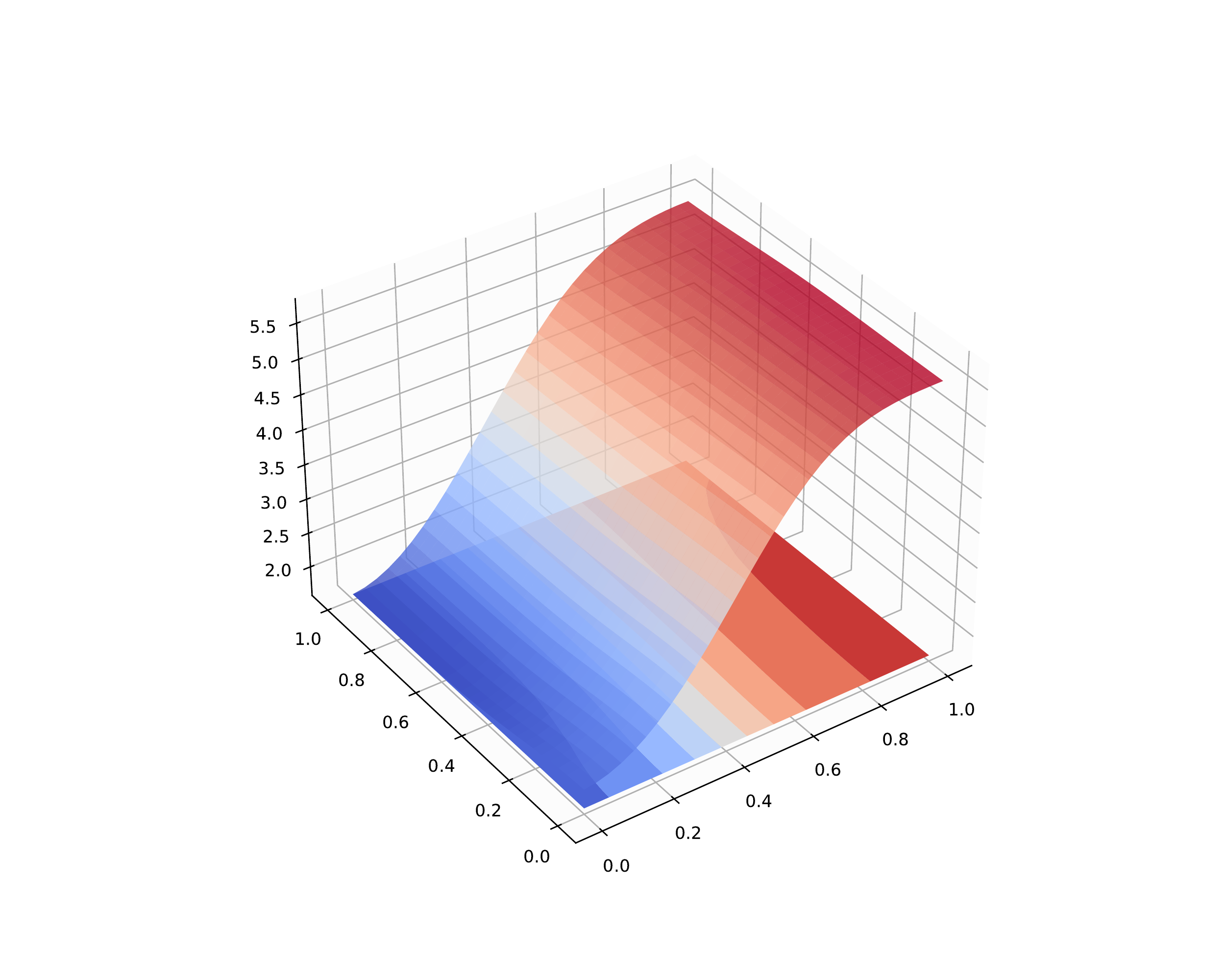}}
	\caption{Ablations on attack generation through the lens of loss landscape. We plot the cross entropy loss \textit{w.r.t.} randomly selected input from CIFAR-10 test set, in the form of $x^\prime = x + k\cdot g + v \cdot r$, where respectively $x$ is the origin input, $g$ is the sign of the gradient direction and $r$ (Rademacher) denotes random direction.}
	\label{Fig:ab}
\end{figure}

In this part, we present results to support our claims on the proposed margin loss for attack generation in single-step adversarial training. We first note that the  \cref{Fig:ab} in the second row is slightly flatter and the corresponding contours are relatively unfolding, when compared with the first row. This implies that applying the margin loss for attack generation is indeed more effective than the cross entropy loss. We further note that, the improvement with the use of the proposed adversary is more significant in single-step defenses (\textit{i.e.}, the left and the middle columns) when compared to iterative adversarial training (\textit{i.e.}, the column on the right). This makes our method appealing,  especially when the available budget on the number of steps for attack generation is highly restricted. 

\end{document}